\documentclass[letterpaper]{article} 
\usepackage{aaai25}   
\usepackage{times}  
\usepackage{helvet}  
\usepackage{courier}  
\usepackage[hyphens]{url}  
\usepackage{graphicx} 
\usepackage{natbib}  
\usepackage{caption} 
\usepackage{algorithm}
\usepackage{algorithmic}
\usepackage{todonotes}

\usepackage{newfloat}
\usepackage{listings}
\DeclareCaptionStyle{ruled}{labelfont=normalfont,labelsep=colon,strut=off} 
\floatstyle{ruled}
\newfloat{listing}{tb}{lst}{}
\floatname{listing}{Listing}
\nocopyright

\usepackage{graphicx}
\usepackage{subcaption}
\usepackage{amsmath} 
\usepackage{mathrsfs}
\usepackage{booktabs}
\usepackage{amssymb}
\usepackage{dsfont}
\usepackage{bbold}
\usepackage{booktabs}

\title{MoESD: Mixture of Experts Stable Diffusion to Mitigate Gender Bias}

\author{
    Guorun Wang,
    Lucia Specia\\
}

\affiliations {
    Imperial College London\\
    \texttt{\{g.wang23,l.specia\}@imperial.ac.uk}
}

\usepackage{bibentry}

\begin{document}

\maketitle

\begin{abstract}
Text-to-image models are known to propagate social biases. For example, when prompted to generate images of people in certain professions, these models tend to systematically generate specific genders or ethnicities. In this paper, we show that this bias is already present in the text encoder of the model and introduce a Mixture-of-Experts approach by identifying text-encoded bias in the latent space and then creating a Bias-Identification Gate mechanism. More specifically, we propose MoESD (Mixture of Experts Stable Diffusion) with BiAs (Bias Adapters) to mitigate gender bias in text-to-image models. We also demonstrate that introducing an arbitrary special token to the prompt is essential during the mitigation process. With experiments focusing on gender bias, we show that our approach successfully mitigates gender bias while maintaining image quality.
\end{abstract}

\section{Introduction}
In recent years, large language and vision models such as ChatGPT 4 \cite{OpenAI_GPT4_2023}, DALL\textperiodcentered E \cite{ramesh2021zero} and Stable Diffusion \cite{rombach2022high, podell2024sdxl} have ushered the era of AI-generated content. However, research has shown that these generative models often exhibit social biases during the content generation process, especially in text-to-image generation. For instance,  models tend to generate more images of men when  provided with prompts like ``a successful CEO'', and more images of women when provided with prompts like ``a paralegal'' \cite{friedrich2023FairDiffusion, luccioni2023stable}.  

To mitigate such biases, current methodologies can be broadly categorized into three debiasing paradigms:

\begin{itemize}
  \item Pre-processing the training data to remove bias before training.
  \item Prompt-engineering to restrain the model generation at the deployment stage.
  \item Enforcing fairness on model weights by introducing constraints on the learning objective during training.
\end{itemize}

In pre-processing methods, eliminating bias in the training corpus is a difficult challenge that offers no guarantees \cite{hamidieh2023identifying}.

For prompt-engineering, although leveraging specific gender attributes to instruct the model (e.g., ``a photo of a female plumber'') can help avoid biases \cite{berg2022prompt, friedrich2023FairDiffusion}, it is not a practical solution as it does not fix the model and one cannot guarantee that the model will always be used with bias mitigating prompts.

For changing model weights, the resource-intensive nature of re-training models poses challenges, requiring vast amounts of data (89k text-image pairs in \citet{esposito2023mitigating}) and fine-tuning of the model.

Although recent methods \cite{friedrich2023FairDiffusion, zhang2023inclusive, Parihar_2024_CVPR, gandikota2024unified, wang2024concept} demonstrate that explicitly specifying gender attributes in the prompt and token, or providing gender attribute guidance in the latent space during generation can yield good results, we aim to reduce this manual interventions to the lowest possible level.

We propose a method based on Mixture of Experts for Stable Diffusion with Bias Adapters, which identifies existing biases in pre-trained models and effectively mitigates them without requiring gender-guided prompts during generation. We modify the actual model weights, but in a parameter-efficient manner, requiring only a small amount of data (1.5K) and parameters (5.6\%) to be updated. Our contributions can be summarized as follows:

\begin{itemize}
  \item We measure the gender skew in text to assess gender bias in embeddings and propose a Bias Identification Gate mechanism.
  \item We introduce Mixture of Experts (MoE) to Stable Diffusion and fine-tune the Bias Adapters (BiAs) to effectively mitigate identified gender biases.
  \item We investigate the use of special tokens to aid the BiAs in better understanding the biased data and demonstrate that it is essential for mitigation, even for challenging professions.
  \item We show that the proposed method successfully mitigates the gender bias in Stable Diffusion generation while maintaining image quality.
  \item We conducted additional experiments on skin tone biases to show the generalisation ability of our method to other biases.
\end{itemize}

\section{Related Work}

Biases in multimodal settings have attracted increasing attention. Several studies have evaluated multimodal models and found that they inherit and propagate various biases \cite{agarwal2021evaluating, Cho2023DallEval, bansal-etal-2022-well}. For example, \citet{bansal-etal-2022-well} introduced the ENTIGEN framework to evaluate and mitigate gender bias.
Prompt and token learning with attributes, as well as attribute editing in the latent space, have been widely utilized in vision-language models and generative models \cite{berg2022prompt}. Specifically, Fair Diffusion \cite{friedrich2023FairDiffusion} explicitly inserts gender attributes into the SEGA model \cite{brack2023Sega}. ITI-GEN \cite{zhang2023inclusive}, similar to Fair Diffusion, learns a set of prompt embeddings to represent gender attribute categories. Balancing Act \cite{Parihar_2024_CVPR} trains a classifier to provide gender attribute guidance in the latent space. UCE \cite{gandikota2024unified} directly edits concepts based on phrases and attributes, while Concept Algebra \cite{wang2024concept} manipulates and combines concepts in the latent space using algebraic operations.
Subsequently, image generation is guided towards a fairer outcome through manual semantic editing in the latent space of biased concepts. 
Although these methods can yield better fairness results in the generated images, our focus is on addressing the biases inherent in the model itself. Therefore, our method learns to modify the model itself and exploits arbitrary tokens and embeddings without gender guidance in the generation process.

Approaches to offset bias representation in the model are also increasingly popular: \citet{seth2023dear} employs additive residual image representations to mitigate biased representations, while 
\citet{esposito2023mitigating} and Fairness Fine-tuning \cite{shen2024finetuning} focuses on fine-tuning the model to achieve fairness. 
DebiasVL \cite{chuang2023debiasing} proposed a method to project out biased directions in text embeddings to create fair generative models leveraging positive pairs of prompts to debias embeddings. By generating embeddings of prompts such as ``a photo of a [class name] with [spurious attribute]'', a calibrated projection matrix, as shown in Equations \ref{eq:mit1}, is optimized. After projection, the embedding should only contain information about the ``[class name]'' with no spurious information (e.g., gender).

Equation \ref{eq:mit1} illustrates the regularization of the difference between the projected embeddings of the set of positive pairs $S$, where $(z_i, z_j)$ represents the embedding of prompt pair $(i, j)$ in $S$, which describes the same class but with different spurious attributes (e.g., gender). The loss function encourages the linear projection $P$ to be invariant to the difference between the spurious attributes (details in Appendix \ref{ap:chuang}). $P_0$ is the encoding process of the pre-trained Stable Diffusion.

\begin{equation}
\label{eq:mit1}
\min_{P} \left\| P - P_0 \right\|^2 + \frac{\lambda}{\left| S \right|} \sum_{(i,j) \in S} \left\| Pz_i - Pz_j \right\|^2
\end{equation}

In our work, we utilize the Projection $P$ from Equation \ref{eq:mit1} to calculate gender skew from the prompt and use the information to better fine-tune the model to offset bias.

\section{Method}

In this section, we describe our method. First, we explore the projection matrix from DebiasVL (Equation \ref{eq:mit1}) to assess inheriting gender biases in Stable Diffusion. Second, we introduce the Mixture of Experts (MoE) to Stable Diffusion and add fine-tuned BiAs (experts), which combine bias identification gates and special tokens to aid in understanding the biased data. This results in our MoESD-BiAs approach.

\subsection{Identifying Biases in Stable Diffusion}

Stable Diffusion, the text-to-image model in our work, generates images from text by switching the diffusion process from pixel space to latent space to generate images. Given an image $x \in \mathbb{R}^{H \times W \times 3}$ in RGB space, the encoder $\mathcal{E}$ encodes $x$ into a latent representation $z = \mathcal{E}(x)$, and the decoder $D$ reconstructs the image from the latent, giving $x' = D(z) = D(\mathcal{E}(x))$, where $z \in \mathbb{R}^{h \times w \times c}$. A domain-specific encoder $\tau_\theta$  projects $y$ (language prompt) to an intermediate representation $\tau_\theta(y) \in \mathbb{R}^{M \times d_\tau}$, which is then mapped to the intermediate layers of the U-Net \cite{ronneberger2015u} via a cross-attention layer implementing to condition the latent $z$. Based on image-conditioning pairs, the conditional LDM is learned via Equation \ref{eq:sd}.

\begin{equation}
\label{eq:sd}
L_{LDM} := \mathrm{E}_{\mathcal{E}(x), y, \epsilon\sim\mathcal{N}(0,1), t} \left[ \left\| \epsilon - \epsilon_\theta (z_t, t, \tau_\theta(y)) \right\|_2^2 \right]
\end{equation}

However when employing prompts like ``a photo of a [occupation]'', the model has been shown to exhibit significant gender bias with stereotypes \cite{friedrich2023FairDiffusion, chuang2023debiasing}, which manifests throughout the entire process.

While research efforts have been dedicated to addressing biases in the iterative image denoising process within U-Net, the convolutional network component of Stable Diffusion \cite{esposito2023mitigating}, our investigation has revealed a distinct bias emerging after the text embedding stage when utilizing the text encoder of different versions of Stable Diffusion. This means that bias is amplified throughout the process. 

We show a visualization of the dimensionality-reduced embeddings encoded by the text encoder of two versions of Stable Diffusion\footnote{\url{https://huggingface.co/runwayml/stable-diffusion-v1-5}} \footnote{\url{https://huggingface.co/stabilityai/stable-diffusion-2-1-base}} using T-SNE, with prompts  like ``\textit{The photo of the face of a [occupation]}'', where the [occupation] are from the top 8 male-biased and top 8 female-biased occupations from Fair Diffusion \footnote{\url{https://github.com/ml-research/Fair-Diffusion/blob/main/results_fairface_generated_1-5.txt}}, (details in Appendix \ref{ap:TSNE}). 
There is a clear boundary between two gender-biases occupation encoded embeddings in both versions of Stable Diffusion. In other words, embeddings show gender biases regardless of whether they are encoded by CLIP or OpenCLIP \cite{radford2021learning, schuhmann2022laionb, cherti2023reproducible}, indicating that the text encoder already contains gender biases, which then conditions the U-Net. Thus, it is not enough to mitigate biases in the latent space of the U-Net, we need to address the text-encoded bias as well.

\subsection{Measuring Text-Encoded Bias}
\paragraph{Issues of  Projecting Prompt Embeddings for Fairness}

Although DebiasVL claims great success in measuring bias in text embeddings and projecting it to achieve fairness, we have identified an issue. When attempting to switch the original prompt into another direction to achieve balance, 
the projection may take it into another direction that can introduce other biases. We identified failure cases in gender mitigation using DebiasVL's method, which resulted in only male faces being generated, details in the Appendix \ref{ap:failure_mit}.


\paragraph{Reframing the Projection to Assess Gender Bias}
We conduct zero-shot and unsupervised classification on embeddings, using prompts like ``\textit{The photo of the face of a [occupation]}'' and a pre-trained Stable Diffusion text encoder. Our goal is to measure gender bias within these prompts and their embeddings. Altering the weights of the pre-trained text encoder is risky, as it could lead to the model forgetting information that is not related to gender, which has been learned from extensive data, so the Stable Diffusion text encoder is frozen. Another challenge is the lack of clear labels, as we only have prompts, embeddings, and statistics for each occupation generated by Fair Diffusion. These statistics are based on 250 images from the original Stable Diffusion and cannot be quantitatively analyzed or treated as a simple binary classification task due to the small sample size. Moreover, the embeddings are subject to change based on the hyperparameters of the pre-trained model, meaning the statistics can only serve as a benchmark for testing our approach rather than for model learning. Therefore, our analysis relies solely on prompts and model weights.
As shown in Equation \ref{eq:mit1}, $z$ is the prompt embedding, and $P$ is the projection process. Prompts $z_0$ applying  \text{Calibration Matrix}  can be written as $Pz_0=P_0 \left( I + \frac{\lambda}{\left| S \right|} \sum_{(i,j) \in S} (z_i - z_j)(z_i - z_j)^T \right)^{-1}z_0$, derived from Equation \ref{eq:mit1}. Thus, we can solve for $P$.

Here we first define the similarity between two prompts before and after applying the \text{Calibration Matrix}: 

\begin{align}
\Delta  S(z_0, z_t, P_0, P) &= \bigl| \text{Similarity}(P_0z_0,P_0z_t) \notag \\
&\quad - \text{Similarity}(Pz_0,P_0z_t) \bigr|
\end{align}

We then calculate the gender skew for prompt embedding $z_0$ as 
\begin{align}
\label{eq:gender}
\mathscr{G}(z_0) &= \Delta  S(z_0, z_{male}, P_0, P) \notag \\
&\quad - \Delta  S(z_0, z_{female}, P_0, P) 
\end{align}

$z_{male}$ is the prompt with `male' forcement, while $z_{female}$ is the prompt with `female' forcement. For instance, $z_0$ is `a photo of the face of a nurse',  $z_{male}$ is `a photo of the face of a male nurse', and $z_{female}$ is `a photo of the face of a female nurse'.

We assume the gender skew is male when $\mathscr{G}(z_0) > 0 $ while the gender skew is female $\mathscr{G}(z_0) < 0 $. If $\Delta  S(z_0, z_t, P_0, P) $ is larger, then the disparity between two prompts before and after applying the \text{Calibration Matrix} is also larger. This results in a greater divergence between the original embedding direction and the  projected embeddings. Consequently, the strength of the redirected embeddings is increased, indicating a stronger gender preference before redirecting. In essence, if the redirection strength is substantial, it indicates a more forceful adjustment away from the original gender direction, which means a higher gender skew.

\subsection{Defining the Bias Identification Gate}
\label{sec:gate}

We utilize the Pearson Correlation Coefficient to measure similarity and check statistical approximation of genders in Fair Diffusion occupations, with further details provided in Appendix \ref{ap:occupation}. If the male count for a particular occupation surpasses half of the total, we assume this occupation is male-skewed; otherwise, it is female-skewed. To assess the effectiveness of our approach when designing the Bias Identification Gate, we compare the frequency-based labels with our calculations derived from Equation \ref{eq:gender} to compute accuracy. We achieve an accuracy rate of 79\%, indicating that our bias measurement approximation, achieved through task reframing, aligns with both intuition and statistical trends. 
The current accuracy is considered sufficient for mitigation since it is not a binary classification task. Additionally, we conducted extra experiments on skin tone after submission, achieving a 68\% accuracy with the gating mechanism, and we can also be able to mitigate the bias.

The accuracy is lower when it comes to occupations such as insurance agents (details in Appendix \ref{ap:gender gate}), as these are mostly likely at the boundary between male and female. In Table \ref{tab:fairness}, we  show mitigation of biases even when classifications of these challenging occupations are incorrect.

\subsection{BiAs Experts: Bias Adapter Experts}
\label{sec:experts}

The next step is to set up the experts for the MoE approach. For that, we implement a way to personalize Stable Diffusion to guide our model towards fairness.

The intuition is to guide the model to generate more female images when the original embedding exhibits the male skew and vice versa. To achieve this, we divide our training text-image pairs into two groups: male and female. We then fine-tune the model separately to generate male-biased and female-biased experts. When the gate is activated, it guides the experts according to the following rules: male skew for female-biased experts and female skew for male-biased experts. This process is called \textbf{bias fine-tuning}.

\begin{figure}[htbp]
  \centering
  \includegraphics[width=0.35\textwidth]{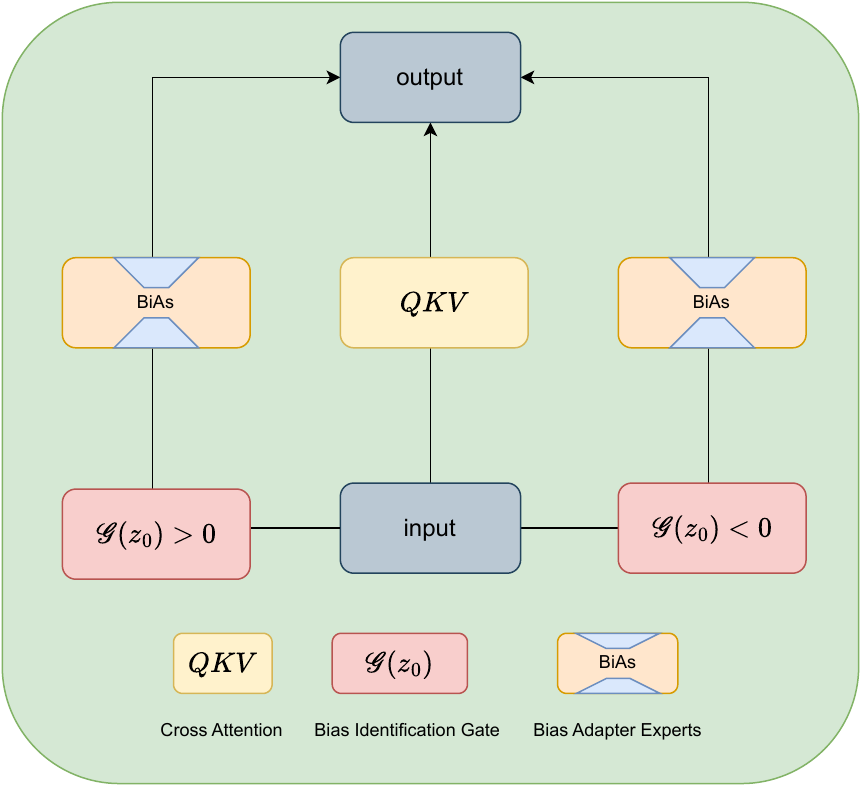}
  \caption{\textbf{The Architecture of BiAs.} We apply BiAs to the cross-attention of the U-Net. In the figure, the left BiAs represents the female expert, the right one, the male expert, and the middle one, the original cross-attention. The conditional information is processed by the Bias Identification Gate to determine which experts to choose - either male, female, or none of them. In the end, the chosen BiAs will process the input and be added together with the original cross-attention. The detailed BiAs architecture includes BiAs of the Q, K, V, and output matrices (for simplicity, we omit them in the figure). The parameters in cross-attention are frozen during the parameter-efficient fine-tuning process.}
  \label{fig:BiAs}
\end{figure}

Adapters \cite{hu2022lora,houlsby2019parameter,karimi2021compacter} provide a method to freeze the model and introduce a new, trainable weight matrix, which significantly reduces both the time and memory required for training. 

We incorporate this method into creating bias experts, making the \textbf{bias fine-tuning} parameter-efficient. We freeze the parameters in the U-Net and add adapters on the cross-attention layers of the U-Net, as illustrated in Figure \ref{fig:BiAs}. We initialize the adapters following the principles of LoRA \cite{hu2022lora}: the weight parameters of the first matrix $W_{\text{down}}'$ are determined by a Gaussian function, while the parameters of the second matrix $W_{\text{up}}'$ are initialized as a zero matrix. Consequently, the added pathway is initially zero during training and does not impact the result, resembling the original output.

Since the adapters are randomly initialized, they do not inherit biases from Stable Diffusion fine-tuning, which makes them more effective in achieving fairness, as detailed in \textbf{Section \ref{sec:fair_adapter}}. 
By using adapters as the bias experts, we achieve better fairness scores and reduce the trainable ratio to only 5.6\%.

For \textbf{bias fine-tuning}, we generate biased images with specific male or female characteristics so that the bias expert will learn stereotypes from the data. We employ a technique from Dreambooth \cite{ruiz2023dreambooth}, utilizing a \textbf{special token} to help BiAs to better understand the biased data. 
Dreambooth adds a special token, a unique identifier, to the prompt, and uses a few fine-tuning images of a subject as input to ``personalize'' the model. In our work, we use the special token to ``personalize'' the adapters of the biased information.

In theory, the special token should not impact the results, as it is only used to remind the experts about the bias. However, due to the context differences of prompt embeddings, minor differences in results might occur when different special tokens are used for different experts and prompts.

For fine-tuning data, we generate our training image-text pairs using Stable Diffusion-XL with the following prompt: \textbf{``A [gender] + [race] + [occupation].''} The prompts generate 1530 images with different genders, races and, occupations to ensure variety, which then fine-tune the model using the special token. The results demonstrate that the special token yields better performance than ordinary fine-tuning, as detailed in Section \ref{sec:fair_adapter}.

Our bias fine-tuning can be formalised in the following optimization process:

\begin{equation}
L_{bias} :=  \mathrm{E}_{\mathcal{E}(x), y, \epsilon\sim\mathcal{N}(0,1), t} \left[ \left\| \epsilon - \epsilon_\theta (z_t, t, \tau_\theta(s)) \right\|_2^2 \right]
\end{equation}

\noindent{where $s$ is the prompt with the special token.}

What distinguishes our work from that of \citet{esposito2023mitigating} is that we only need to fine-tune a small set of data, rather than 89K prompts and 89K images, and fine-tune a small ratio of parameters, making it more training-efficient. 

What distinguishes our work from methods such as Fair Diffusion is that we do not manually engage with gender attributes in the prompt or latent space; in other words, we do not introduce human intervention. The special token we add can be any token. Moreover, even without the special token, our approach still mitigates gender bias.

\subsection{Mixture of Experts}
\label{sec:MoE}
MoE (Mixture of Experts) \cite{jacobs1991adaptive, eigen2013learning} is an ensemble learning technique that improves performance by weighting the predictions of different experts through gate mechanisms.

In a classic MoE system, each expert is independent and demonstrates high performance within their area of expertise. A gating mechanism is learned to adjust the weight assigned to each expert based on the input data. 
In contrast, in our approach, the gates and experts, defined in Section \ref{sec:gate} and \ref{sec:experts}, are not trained. We utilize pre-trained models, which can be flexibly replaced by other pre-trained models.

\begin{figure*}
    \centering
    \includegraphics[width=0.7\linewidth]{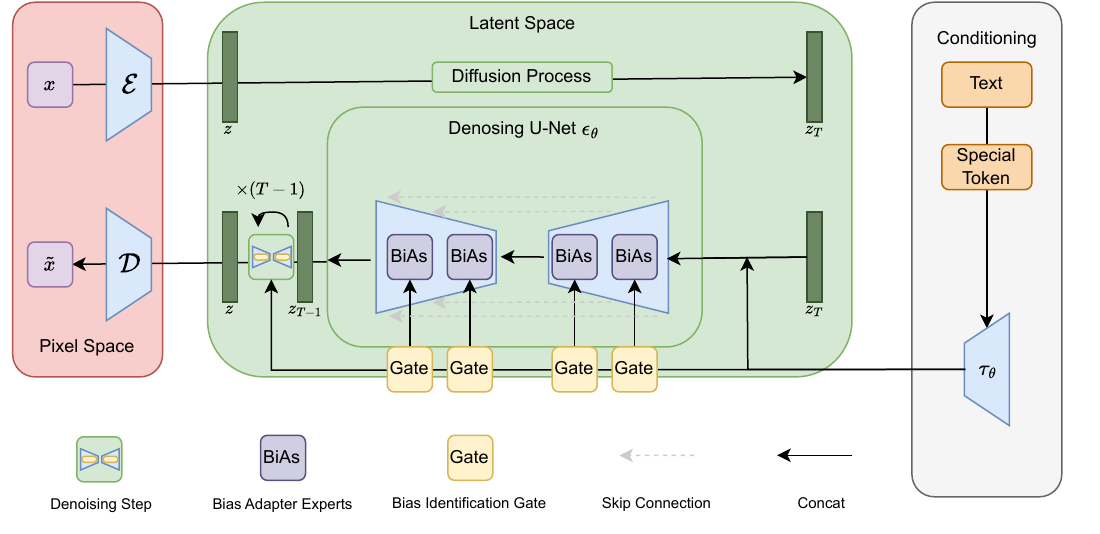}
    \caption{\textbf{Architecture of our MoE system.} We modify  \citet{rombach2022high} by adding our Bias Identification Gate and Bias Adapter Experts. See Section \ref{sec:MoE} for details.}
    \label{fig:MoE}
\end{figure*}

We show our MoE architecture in Figure \ref{fig:MoE}. 
What makes our gating mechanism different from traditional MoE is that we do not learn it from the data, and for the experts, we utilize pre-trained models.

To summarise, our approach is as follows: we first conduct zero-shot and unsupervised classification. We do not use the labels (the gender statistics of the 250 images for each occupation generated by the vanilla Stable Diffusion, provided by Fair Diffusion) to train the model, but only for hyperparameter selection. In other words, we only use the prompt and pre-trained CLIP encoder itself. At inference, the model takes the original prompt and special token as inputs. It judges the gender skew the prompt shows, with the male skew leading to a higher probability of calling the female expert, and vice versa. We show our mitigation examples in Appendix \ref{ap:show}.

\section{Experiments}

\subsection{Setup}

We first use the statistical data of different occupations generated by Fair Diffusion (Stable Diffusion version 1.5) to select the best hyperparameters for both versions of Stable Diffusion. If the general statistical count shows more males than females, we assume that it has a male skew, and vice versa. We found that 4000 is the optimal hyperparameter $\lambda$ in Equation \ref{eq:mit1} and the subsequent derivation for version 1.5, and 100 is optimal for version 2.1. 
There is a large difference in the best hyperparameter between the two versions, which we attribute to the differences in the text encoder between the two versions of Stable Diffusion, which introduces different embedding gender biases in the latent space.

Next, we use Stable Diffusion XL to generate fine-tuning images data and then fine-tune our MoESD with BiAs experts by adding a special token to the original prompts (set to ``sks'' as in \citet{ruiz2023dreambooth}). The prompt is ``A photo of the sks face of the [occupation].''

After fine-tuning, we proceed to the inference steps: There are three experts in the system: one original, one male BiAs expert, and one female BiAs expert. When inputting a prompt, we add the same special token (``sks'') as in the fine-tuning stages. The system judges the skew, and if the skew is male, we allocate 10\% to the male expert, 50\% to the female expert, and 40\% to the original expert. Conversely, if the skew is female, we allocate 10\% to the female expert, 50\% to the male expert, and 40\% to the original expert. These are the optimal hyperparameters for conservative mitigation within a limited search range in our experiments. We do not want the BiAs to completely change the original weights, as our Bias Identification Gate mechanism is not absolutely accurate.
In this manner, we generate 100 images based on the same prompt for each of the 153 occupations.

We then employ BLIP2 \cite{li2023blip} VQA task to conduct fairness evaluation, evaluation method of \citet{esposito2023mitigating}, and Laion-aesthetic linear classifier \cite{LAION-AESTHETICS} to perform aesthetic evaluation, as we discuss in the following sections. We also employ human evaluation for image-description relevance evaluation. 

In our method, we do not explicitly provide any gender information in the prompt, token, or gender guidance in the latent space. Therefore, we do not compare our method with those that include gender hints and guidance.

\subsection{Fairness Evaluation}

\subsubsection{Metric}

Stable Diffusion learns a conditional distribution \( \hat{P}(X|Z=z) \), where \( z \) represents the embedding of the prompt, and the biased nature of the dataset used to train the generative model and its architecture can impact the distribution \( \hat{P} \). To quantify the bias in generative models, recent studies \cite{choi2020fair, teo2024measuring, chuang2023debiasing} propose using statistical parity. Specifically, given a classifier \( h: X \rightarrow A \) for the gender attribute, the discrepancy of the generative distribution \( \hat{P} \) can be defined as two metrics: the empirical and uniform distributions.  

We use the metric in the evaluation of the images generated by Stable Diffusion, where \( X \) is the set of images for each occupation \( o \) from Fair Diffusion, classifier \( h \) is the BLIP2 model with prompts to predict gender. \( A \) is a set containing ``male'' and ``female'', and the expectation is estimated with empirical samples. We assume the model to be fairer when the empirical distributions are closer to uniform distributions, since a fair model minimizes the discrepancy by ensuring that each attribute \( a \in A \) has a similar probability (uniformly distributed).

Equation \ref{eq:metric2}  incorporates ideal fair expectations into the estimation of empirical samples for each occupation. While this metric may not consider the overall gender ratio and may sometimes favor a higher proportion of males/females in all occupations, it aims to detect gender bias within each occupation. The standard deviation helps identify occupations that deviate significantly from the fair boundary. 
\begin{equation}
    \frac{1}{\left|O\right|} \sum_{a \in A, x \in O} \left| \mathrm{E}_{x \sim \hat{P}} \left[ \mathbb{1}_{h(x)=a} \right] - \frac{1}{|A|} \right|
    \label{eq:metric2}
\end{equation}

In the context of binary gender, Equation \ref{eq:metric2} for each attribute \( a \in A \), male or female, is the same. Therefore, for simplicity, we report the male attribute in the results. 

\subsubsection{Results}

We compare our method to the original Stable Diffusion, DebiasVL, Ethical Intervention, and Fairness Fine-tuning across versions 1.5 and 2.1. Due to the variability in results related to hyperparameters reported by DebiasVL, we select the best results from different hyperparameters for their method in both versions: 0.05 for version 1.5 and 500 for version 2.1, as reported in their paper. For Ethical Intervention, we use the prompt intervention: ``if all individuals can be a \textit{[occupation]} irrespective of their gender.'' For Fairness Fine-tuning, we use the provided pre-trained weights of version 1.5 and report three different fine-tuning methods: fine-tuning the text encoder (TE), fine-tuning the U-Net, and fine-tuning both of them. MoESD-BiAs (with special token) performs better than all others.
The results are shown in Table \ref{tab:fairness}. Below we summarise our main observations:

\begin{table*}[ht!]
\centering
\small
\begin{tabular}{@{}llcccccccccc@{}}
\toprule
\multicolumn{2}{c}{\textbf{Model Version}} & \multicolumn{2}{c}{\textbf{Stable Diffusion 1.5}} & \multicolumn{2}{c}{\textbf{Stable Diffusion 2.1}} \\
\cmidrule(r){1-2} \cmidrule(lr){3-4} \cmidrule(l){5-6}
\textbf{} & \textbf{Method} & \textbf{Fairness Score} $\downarrow$ & \textbf{STD} $\downarrow$ &  \textbf{Fairness Score} $\downarrow$ & \textbf{STD} $\downarrow$ \\
\midrule
Benchmarks & Vanilla & 0.281 & 0.167 & 0.326 & 0.146 \\
& DebiasVL & 0.279 & 0.169 & 0.255 & 0.143  \\
& Ethical Int.  & 0.379 & 0.120 & 0.374 & 0.140  \\

& Fairness Fine-tuning (TE) & 0.239 & 0.132 & - & -  \\
& Fairness Fine-tuning (U-Net) & 0.255 & 0.149 & - & -  \\
& Fairness Fine-tuning (TE+U-Net) & 0.273 & 0.157 & - & -  \\

Ours & MoESD & 0.293 & 0.151 & 0.344 & 0.144 \\
& MoESD-BiAs & 0.169 & 0.124 & 0.274 & 0.147  \\
& MoESD (special token) & 0.141 & 0.1 & 0.147 & 0.108  \\
& MoESD-BiAs (special token) & \textbf{0.135} & \textbf{0.103} & \textbf{0.136} & \textbf{0.107}  \\
\bottomrule
\end{tabular}
\caption{\textbf{Fairness Evaluation - Comparison with Vanilla Stable Diffusion \cite{rombach2022high}, DebiasVL \cite{chuang2023debiasing}, Ethical Intervention \cite{bansal-etal-2022-well}, and Fairness Fine-tuning \cite{shen2024finetuning}.} 
For each method, we show the Fairness Score and Standard Deviation for two versions of Stable Diffusion. Our MoESD-BiAs (special token) yields the best performance in all cases.}
\label{tab:fairness}
\end{table*}

%

\paragraph{\textit{Occupation Robustness}}
Ethical Intervention performs worse than the vanilla model. This is because the method is effective in professions that are not as challenging as those listed in the profession list of Fair Diffusion. Moreover, interventions without gender attributes worsen the fairness results, indicating that excessive intervention makes it difficult for the model to handle challenging occupations.
Our method, through effective fine-tuning on the more difficult occupations, demonstrates robustness in these cases.

\paragraph{\textit{Model Version}}
In most methods, the fairness score for Stable Diffusion Version 2.1 is higher than for Version 1.5, except for Ethical Intervention, which indicates that Version 1.5 performs better in terms of fairness. For other methods, there is a significant difference between different versions. However, our MoESD-BiAs (special token) achieves similarly good performance in both versions.

\paragraph{\textit{Adapter}} \label{sec:fair_adapter}
In our method, BiAs appears to perform better than full fine-tuning regardless of whether we use the special token, which suggests that only a small set of parameters (5\%) are enough to mitigate bias.
Additionally, the results of Fairness Fine-tuning show that Fairness Fine-tuning (TE) and Fairness Fine-tuning (U-Net) perform better than Fairness Fine-tuning (TE+U-Net), indicating that excessive fine-tuning is not always beneficial in mitigating gender bias.
We attribute it to the fact that adapters allow more targeted parameter updates, reducing the risk of gradient vanishing or exploding and preventing the catastrophic forgetting of general knowledge from well pre-trained Stable Diffusion weights. Moreover, due to random initialization, our adapters are more effective in mitigating bias in relatively small data and a few fine-tuning steps.

\paragraph{\textit{Special Token}} 
The arbitrary special token contributes substantially to mitigating bias since it helps the model (bias experts) to remember the fine-tuning process, which is especially effective for the randomly initialized adapters. We can see the MoESD method performs worse than the vanilla method. We attribute it to the fact that small fine-tuning cannot impact biases of a large base model. However, the special token can mitigate this, as seen from the MoESD (special token) results.

\subsection{Aesthetic Evaluation}

\subsubsection{Metric}

It should be noted that due to differences in training data, the style and quality of generated images vary. 
Since our training dataset is AI-generated instead of using real images, we noticed that some of our images contain blurred faces. Therefore, we need to find a way to evaluate the image quality. However, due to the synthetic nature of the training dataset and generated content, we cannot use the Fr\'{e}chet  Inception Distance (FID) \cite{heusel2017gans} for evaluation.

Simulacra Aesthetic Captions - SAC \cite{pressmancrowson2022} is a dataset consisting of over 238,000 synthetic images generated with AI models, from over forty thousand user submitted prompts. Users rate the images on their aesthetic value from 1 to 10,  when they were asked ``How much do you like this image on a scale from 1 to 10?''. 
LAION-Aesthetics \cite{LAION-AESTHETICS} trains a linear model on 5000 image-rating pairs from the SAC dataset, which can predict a numeric aesthetic score in 1-10.
We utilize this linear model to evaluate the image sets of all methods and compare their aesthetic scores.






\subsubsection{Results}

To better evaluate our fine-tuning results, we compare all the methods on the aesthetic level to determine whether fine-tuning affects abilities other than fairness. The results in the Table \ref{tab:aesthetic}  show that our method maintain the image quality according to this metric.

\begin{table*}[ht!]
\centering
\small
\begin{tabular}{@{}llcccccccccc@{}}
\toprule
\multicolumn{2}{c}{\textbf{Model Version}} & \multicolumn{1}{c}{\textbf{Stable Diffusion 1.5}} & \multicolumn{1}{c}{\textbf{Stable Diffusion 2.1}} \\
\cmidrule(r){1-2} \cmidrule(lr){3-4} \cmidrule(l){5-6}
\textbf{} & \textbf{Method} & \textbf{Aesthetic Score} $\uparrow$ &  \textbf{Aesthetic Score} $\uparrow$ \\
\midrule
Benchmarks & Vanilla & 5.131 & 5.298  \\
& DebiasVL & 5.132 & 5.285   \\
& Ethical Int.  & 5.134 & 5.286  \\

& Fairness Fine-tuning (TE) & 5.022 & -  \\
& Fairness Fine-tuning (U-Net) & 5.024 & -  \\
& Fairness Fine-tuning (TE+U-Net) & 5.067 & -  \\

Ours & MoESD & \textbf{5.423} & 5.439 \\
& MoESD-BiAs & 5.406 & \textbf{5.483} \\
& MoESD (special token) & 5.348 & 5.452  \\
& MoESD-BiAs (special token) & 5.290 & 5.455  \\
\bottomrule
\end{tabular}
\caption{\textbf{Aesthetic Evaluation - Comparison with Vanilla Stable Diffusion \cite{rombach2022high} , DebiasVL \cite{chuang2023debiasing}, Ethical Intervention \cite{bansal-etal-2022-well}, and Fairness Fine-tuning \cite{shen2024finetuning}.} For each method, we show the Aesthetic Score for two versions of Stable Diffusion. Our method yields the best performance in both versions.}
\label{tab:aesthetic}
\end{table*}

Compared to the vanilla approach, MoESD-BiAs (special token) has a higher aesthetics score, even though it may blurs faces at times. We speculate that the higher aesthetics score may be because of images with more colorful and complex backgrounds, which are favored by the aesthetic classifier. Additionally, we also observed that the vanilla method sometimes produces images with words, which receive very low aesthetics scores, as also noted in the Laion-aesthetic example\footnote{\url{http://captions.christoph-schuhmann.de/aesthetic_viz_laion_sac+logos+ava1-l14-linearMSE-en-2.37B.html}}. Finally, the score for version 2.1 is higher than for version 1.5, indicating that version 2.1 has better aesthetic recognition during pre-training. 

Our method with the special token sometimes performs worse than our method without the special token, suggesting that the special token may be occasionally harmful to aesthetics. 



\subsection{Image-Description Relevance Evaluation}

We also launched an additional human evaluation on the quality of our generated images to further understand the results in the previous section. We selected the same 30 images generated by us from the aesthetic evaluation to assess how well each image matches its description (the prompt given to Stable Diffusion). The instruction is ``\textit{How well does the image matches the image description. 10 stands for fully relevant and 0 stands for not relevant. 
Image description: A photo of the face of the [occupation name].}''

Out of the 30 images, 10 images do not activate the bias experts, in other words, there are no changes to the vanilla weight, while 20 images activate the bias expert.

We then calculated the mean and median relevance scores for each image and conducted an analysis for all 30 images. Details of the survey are listed in Appendix \ref{ap:human}.

In this sanity check, we only aim to prove that our method does not significantly affect the image quality compared to the original. Therefore, we did not include other methods in the human evaluation.
The results are shown in Table \ref{table:human_match}. We observe a small degradation in quality when our bias mitigation method is activated. However, this degradation is acceptable, as the score difference is within a range of 1 (in 10).

\begin{table}[ht]
\centering
\small
\begin{tabular}{lcc}
\hline
\textbf{Method} & \textbf{Mean} $\uparrow$ & \textbf{Median} $\uparrow$ \\
\hline
Non-Activated BiAs (10)  & 7.17 & 7.5   \\
Activated BiAs (20) & 6.48 & 6.8 \\
Aggregated (30) & 6.71 & 7.03 \\
\hline
\end{tabular}
\caption{\textbf{Mean and Median of Human Evaluation.} 
The number in parentheses represents the number of images.}
\label{table:human_match}
\end{table}

\subsection{Debiasing for Skin Tone}

Here we applied our method for a different bias mitigation task: 
bias of race, as defined by skin color. We evaluate our method on the skin tone task of \citet{esposito2023mitigating}, which is simpler than using complex race categories. 
The task is similar to debiasing for gender: it uses the same profession prompts to generate images, but evaluates the images at the skin tone level. Evaluation details can be found in the Appendix \ref{ap:VQA}.
Our results indicate that while the performance of the bias identification gate does not match the accuracy levels seen in other common binary classification tasks, it still plays a crucial role in mitigating gender and skin tone bias. We show our mitigation result in Table \ref{table:skin} compared with the Vanilla method in Stable Diffusion v1.5.

\begin{table}[ht]
\centering
\small
\begin{tabular}{lcc}
\hline
\textbf{Method} & \textbf{Fairness Score} $\downarrow$ & \textbf{STD} $\downarrow$ \\
\hline
Vanilla  & 0.209 & 0.134   \\
MoESD-BiAs (special token) & \textbf{0.127} & \textbf{0.106} \\
\hline
\end{tabular}
\caption{\textbf{Effectiveness of our Method on Skin Tone}.}
\label{table:skin}
\end{table}

\section{Conclusions}
We introduced the MoESD-BiAs approach to text-to-image models: an MoE-based method integrated to Stable Diffusion to assess biases in the prompt embeddings, combined with   fine-tuning the BiAs approach with arbitrary tokens to mitigate biases in the model. 
By adjusting the model weights to counteract inherent biases, we generate gender-fair images without requiring specific gender attribute guidance from the prompt, allowing the model to further improve when combined with other methods. 

\clearpage

%


\clearpage
\clearpage
\bibliography{aaai25}

\clearpage
\clearpage

\appendix

\section{Mitigation Visualization}
\label{ap:show}
\begin{figure*}
    \centering
    \includegraphics[width=1\linewidth]{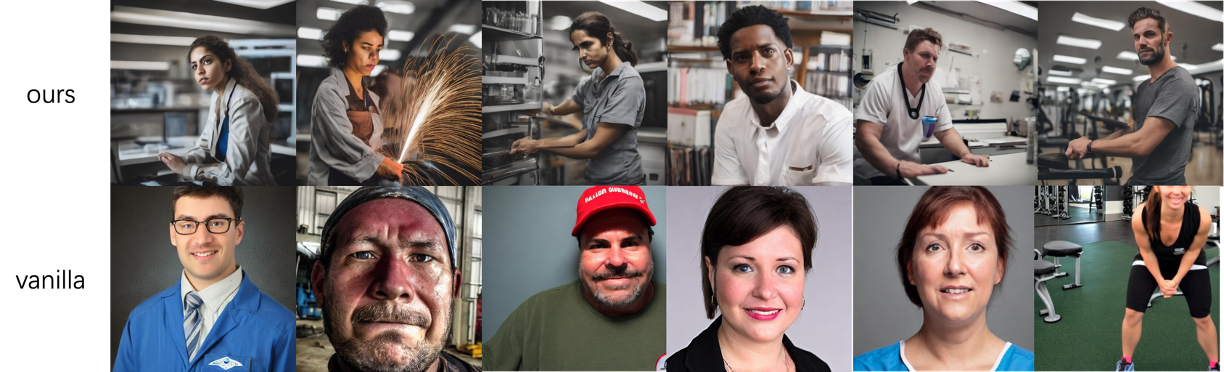}
    \caption{\textbf{Successful Mitigation of Gender Bias.} From left to right are the occupations of aerospace engineer, metal worker, plumber, executive assistant, nurse, and fitness instructor. Each column is generated by the same prompt and seed. The left three are extremely male-biased occupations, and the right three are extremely female-biased occupations. The BiAs Expert successfully leads to a fairer generation. }
    \label{fig:mitigation}
\end{figure*}

We showcase our method with vanilla Stable Diffusion to demonstrate our gender bias mitigation in Figure \ref{fig:mitigation}.

\section{Text-Encoded Bias from Prompts}
\label{ap:TSNE}

\begin{figure*}[htbp]
  \centering
  \begin{subfigure}[b]{0.45\textwidth}
    \includegraphics[width=\textwidth]{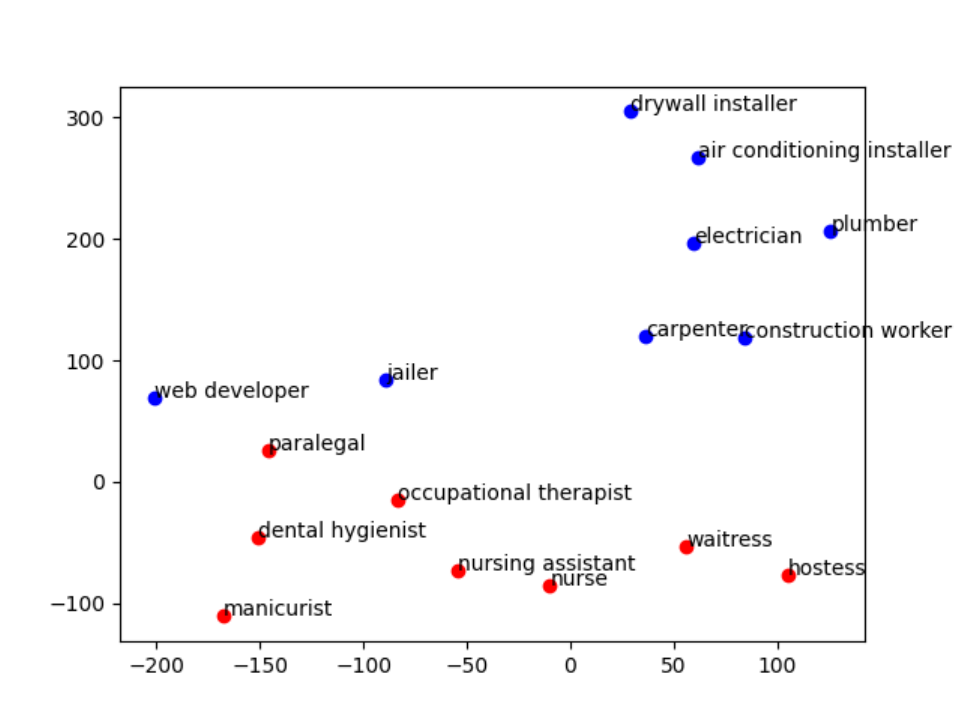}
    \caption{CLIP ViT-L/14 of Stable Diffusion Version 1.5}
    \label{fig:sub1}
  \end{subfigure}
  \hfill
  \begin{subfigure}[b]{0.45\textwidth}
    \includegraphics[width=\textwidth]{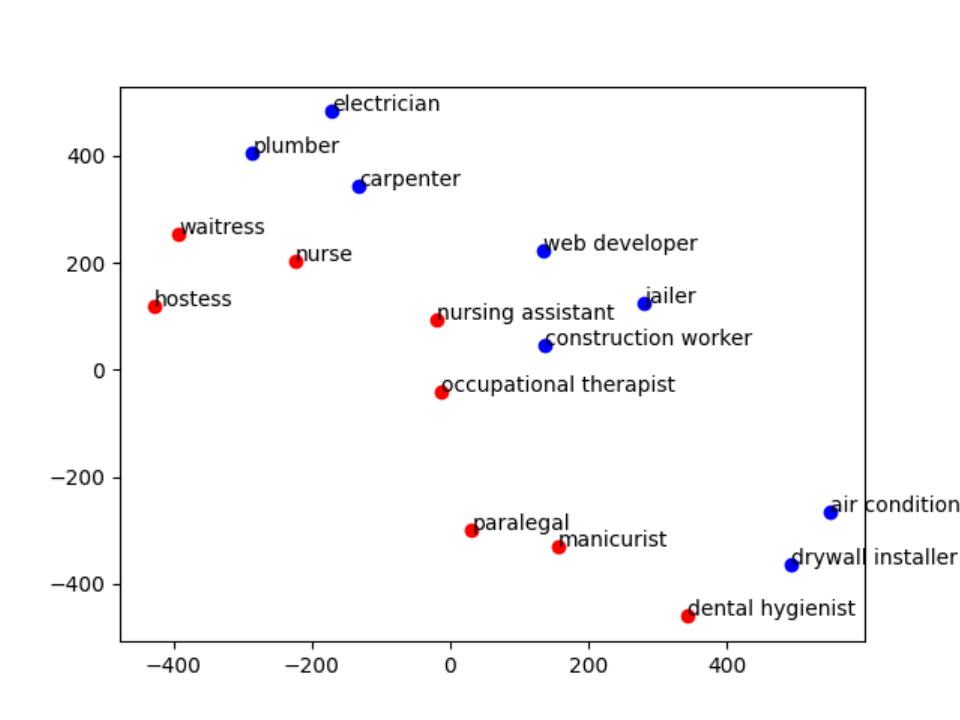}
    \caption{OpenCLIP-ViT/H of Stable Diffusion Version 2.1}
    \label{fig:sub2}
  \end{subfigure}
  \caption{T-Distributed Stochastic Neighbor Embedding (T-SNE) dimension reduction and visualization of prompts encoded by text encoder of Stable Diffusion Version 1.5 and 2.1 \cite{rombach2022high}. There is a clear boundary between two gender-bias occupation  embeddings in both versions of Stable Diffusion.}
  \label{fig:TSNE}
\end{figure*}

We use T-Distributed Stochastic Neighbor Embedding (T-SNE) to visualize the text-encoded bias from prompts, as shown in  Figure \ref{fig:TSNE}. 

\section{Different Methods for Bias-Identification Gate}
We showcase various methods for Bias-Identification Gate, and our approach stands out as the most effective in identifying bias in the prompt according to the Equation \ref{eq:gender}. We directly compare classifications without the intervention of the Calibration Matrix Projection by simply comparing the similarity of $z_0$, $z_{male}$, and $z_{female}$ to determine whether $z_0$ is closer to $z_{male}$ or $z_{female}$. The rule is demonstrated below:

\begin{align}
\label{eq:other}
\mathscr{G}'(z_0) &= Similarity'(z_0, z_{male}) \notag \\
&\quad - Similarity' (z_0, z_{female}) 
\end{align}

We also utilize third-party models (T5  and Sequence Transformer ) as monitoring models to utilize their embedding similarity for classification and compare the results with our method. 

For T5, we perform the QA task for classification, as follows:

\textbf{Question} = ``Answer the following question with `male' or `female'. Is the face more likely to be male or female?''

\textbf{Context} = ``A photo of the face of the '' + [occupation]

For Sentence Transformer, we calculate the similarity between two different set of prompts and perform the classification, selecting the maximum accuracy:

(1) \textbf{Query} = ``A photo of the face of the ''  + [occupation]

\quad \textbf{Docs} = ``A photo of the face of the male'', ``A photo of the face of the female''

(2) \textbf{Query}= [occupation]

\quad \textbf{Docs} = [``male'', ``female'']

Once again, our identification method yields the best results.

\begin{table}[ht]
\centering
\begin{tabular}{lc}
\hline
\textbf{Method} & \textbf{Accuracy} $\uparrow$ \\
\hline
Ours & \textbf{79\%} \\
Cosine Similarity (CLIP) & 72\% \\
Euclidean Distance (CLIP) & 27\%  \\
Manhattan Distance (CLIP) & 28\%  \\
Jaccard Similarity (CLIP) & 43\%  \\
Pearson Correlation Coefficient (CLIP) & 72.5\%  \\
T5 Prompt Classification & 57\%  \\
Sentence Transformer Embedding & 70\%  \\
\hline
\end{tabular}
\caption{
\textbf{Accuracy for Gate-Identification in Different Methods.} The method labeled with ``(CLIP)'' indicates that we use the embeddings from the CLIP encoder obtained from Stable Diffusion.}
\end{table}


\section{More Details of DebiasVL}
\label{ap:chuang}

The detailed proof is shown in the original work. For simplicity in our work, the derivation from Equation \ref{eq:mit1} is presented as follows. Equations \ref{eq:mit2} and \ref{eq:mit3} demonstrate that the calibrated projection matrix and prompt embedding have convenient closed-form solutions.

\begin{equation}
\label{eq:mit2}
P = P_0 \left( I + \frac{\lambda}{\left| S \right|} \sum_{(i,j) \in S} (z_i - z_j)(z_i - z_j)^T \right)^{-1}
\end{equation}

\begin{equation}
\label{eq:mit3}
z^* = \underbrace{\left( I + \frac{\lambda}{|S|} \sum_{(i,j) \in S} (z_i - z_j)(z_i - z_j)^T \right)^{-1}}_{\text{Calibration Matrix}} z_0
\end{equation}

\section{Parameter $\lambda$ for Bias-Identification}

From Equation \ref{eq:gender} and \ref{eq:mit2}, we can observe that the parameter $\lambda$ is crucial for Bias Identification. Therefore, we can deduce that the accuracy of the result is influenced by this parameter. To visualize the results based on $\lambda$, we observe that 4000 is optimal for Stable Diffusion 1.5, while 100 is optimal for Stable Diffusion 2.1.

\begin{figure}[htbp]
  \centering
  \includegraphics[width=0.5\textwidth]{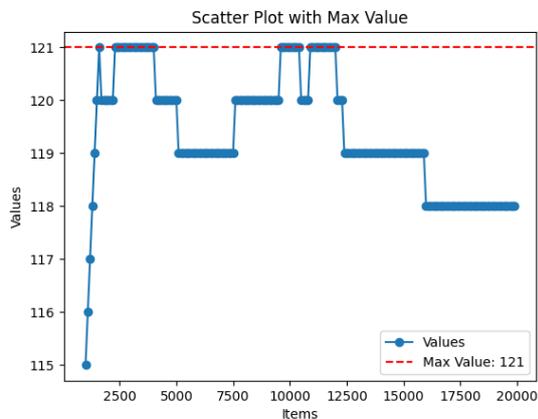}
  \caption{The number of correct predictions for Stable Diffusion 1.5.}
\end{figure}

\begin{figure}[htbp]
  \centering
  \includegraphics[width=0.5\textwidth]{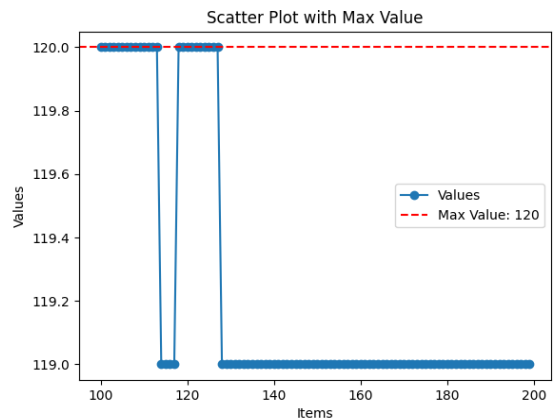}
  \caption{The number of correct predictions for Stable Diffusion 2.1.}
\end{figure}

\section{Defect of DebiasVL} \label{ap:failure_mit}

\begin{figure}[htbp]
  \centering
  \includegraphics[width=0.3\textwidth]{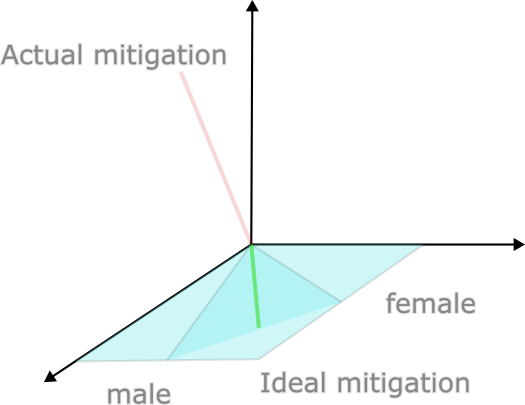}
  \caption{Latent Representation of Prompt Embedding}
  \label{fig:defect}
\end{figure}

As shown in Figure \ref{fig:defect}, the ideal mitigation should lie in the middle of male and female; however, the actual mitigation occurs in another dimension. Although it may have the same meaning when representing male or female, the same meaning may not be neutral and could still contain some gender attributes. 

\begin{figure*}
    \centering
    \includegraphics[width=1\linewidth]{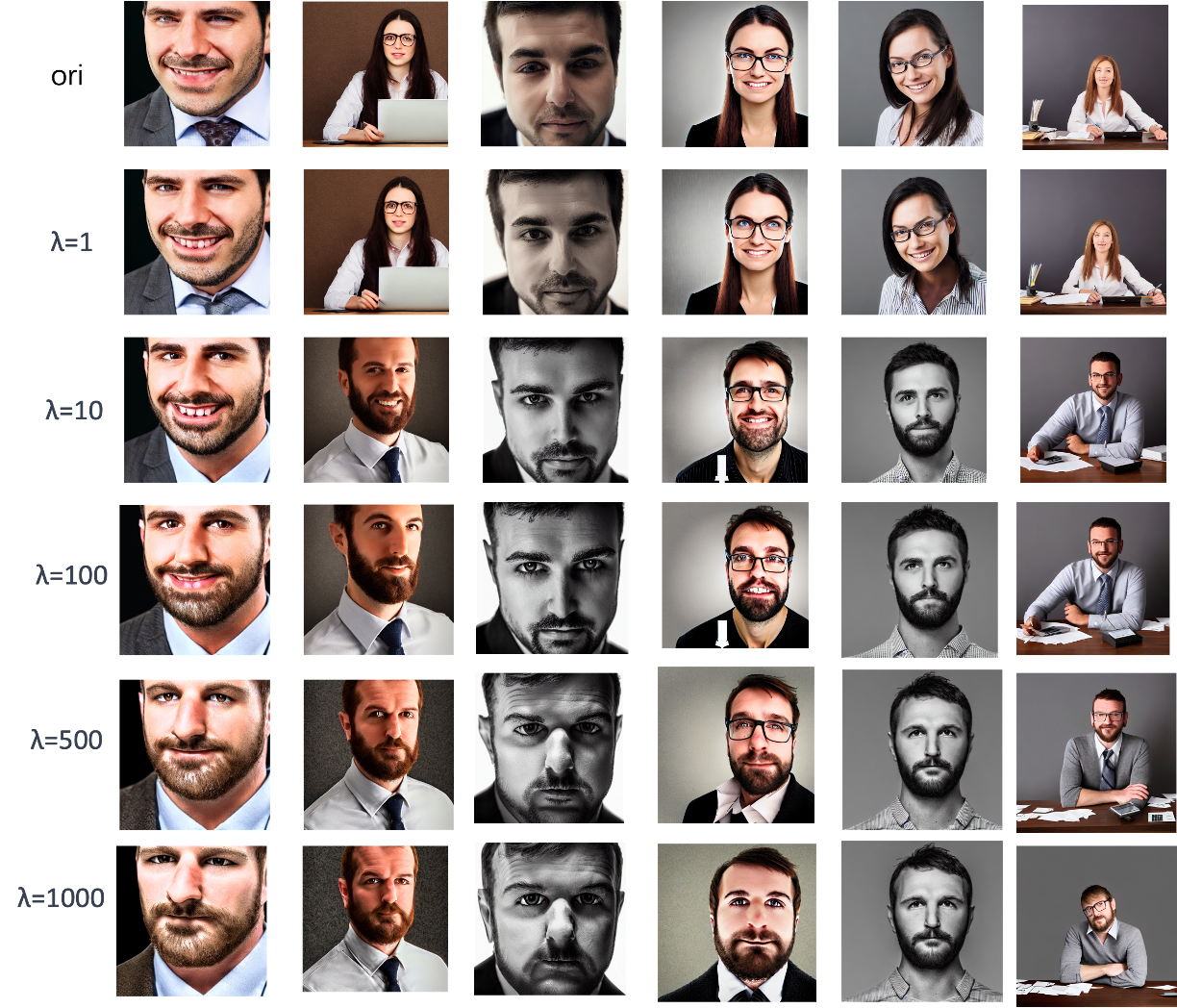}
    \caption{\textbf{Defect of DebiasVL.} The male characteristics become more and more prominent, and in some cases, the gender even switches. }
    \label{fig:defectVis}
\end{figure*}

So DebiasVL exhibits a defect that we have observed: as the parameter $\lambda$ increases, the original characteristics are lost and male characteristics become more and more prominent. We present a different set of images, where each set of images is generated with the same prompt and seeds but with different $\lambda$ to illustrate the ``male guidance'', shown in Figure \ref{fig:defectVis}.

\section{Occupation list} \label{ap:occupation}
We use the occupation list from Fair Diffusion instead of DebiasVL, as we found that the Fair Diffusion one contains more occupations and includes more challenging ones. All occupations are displayed in the Table \ref{tab:occupation} below, where we specifically label the right and wrong occupation predictions for our Bias-Identification method.

\begin{table*}[htbp]
\centering
\begin{tabular}{c}
\toprule
\textbf{Right} \\ \midrule
accountant, aerospace engineer, air conditioning installer, architect, artist, \\
author, bartender, bus driver, butcher, career counselor, carpenter, \\
carpet installer, cashier, ceo, childcare worker, civil engineer, clergy, coach, \\
community manager, compliance officer, computer programmer, \\
computer support specialist, construction worker, cook, correctional officer, \\
customer service representative, dental assistant, dental hygienist, dentist, \\
detective, director, dishwasher, drywall installer, electrical engineer, \\
electrician, engineer, event planner, executive assistant, farmer, \\
fast food worker, financial advisor, financial analyst, financial manager, \\
fitness instructor, groundskeeper, hairdresser, head cook, health technician, \\
host, hostess, housekeeper, industrial engineer, interior designer, interviewer, \\
it specialist, jailer, janitor, language pathologist, librarian, logistician, \\
machinery mechanic, machinist, maid, manager, manicurist, market research analyst, \\
massage therapist, mechanic, mechanical engineer, medical records specialist, \\
mental health counselor, metal worker, mover, musician, network administrator, \\
nurse, nursing assistant, nutritionist, occupational therapist, office clerk, \\
painter, paralegal, payroll clerk, pharmacist, pharmacy technician, pilot, \\
plane mechanic, plumber, police officer, postal worker, printing press operator, \\
programmer, purchasing agent, radiologic technician, receptionist, repair worker, \\
roofer, sales manager, salesperson, scientist, security guard, sheet metal worker, \\
singer, social assistant, social worker, software developer, stocker, taxi driver, \\
teacher, teaching assistant, teller, therapist, tractor operator, truck driver, \\
tutor, waiter, waitress, web developer, welder, wholesale buyer, writer \\ \midrule
\textbf{Wrong} \\ \midrule
aide, baker, claims appraiser, cleaner, clerk, computer systems analyst, \\
courier, credit counselor, data entry keyer, designer, dispatcher, doctor, \\
facilities manager, file clerk, firefighter, graphic designer, insurance agent, \\
inventory clerk, laboratory technician, lawyer, marketing manager, office worker, \\
photographer, physical therapist, producer, psychologist, public relations specialist, \\
real estate broker, school bus driver, supervisor, underwriter, veterinarian \\
\bottomrule
\end{tabular}
\caption{Occupation Set, with right and wrong predictions for our Bias-Identification method.}

\label{tab:occupation}
\end{table*}

\section{BLIP2 VQA for Fairness Evaluation} \label{ap:VQA}
Followed by \citet{esposito2023mitigating}, we use BLIP2 VQA for gender mitigation evaluation and skin tone supplemental experiment.

\subsection{Gender}
For each picture, we use the following question to conduct the VQA evaluation and count the number of males and females in 100 pictures for each occupation.

\textbf{Question}: ``Answer the following question with `male' or `female' or `people not present' only. Is this person on this file male or female?''

For ``people not present'', BLIP2 sometimes gives the answer ``unknown'', so it will not be reflected in the count of ``people not present''. However, it does not matter since we only care about the male and female count for the fairness score.

\subsection{Skin Tone}
For each picture, we use the following question to conduct the VQA evaluation and count the number of skin tone in 100 pictures for each occupation.

\textbf{Question}: `` Answer the following question with `light' or `medium' or `black'
(dark skin) or `people not present' only. What is the skin tone of this person?''

However, the model is not willing to classify skin tone as `medium' or `people not present'; it only categorizes them as either `light' or `dark'. 


\section{Failure cases} 
\subsection{Right and Wrong Case for Bias Identification} \label{ap:gender gate}

As previously presented in the Table \ref{tab:occupation}, which illustrates the right and wrong cases for Bias Identification. We can observe that the wrong cases are most likely to be difficult ones, including very neutral occupations and occupations containing two or more words. 

\subsection{Blurring Generation in Our Method} 

Some of our generated images exhibit blurred faces. We present our human evaluation results for aesthetics, showing that our method was rated the same or better than the vanilla method only in 54.6\% of the cases. This reflects the fact that our images are sometimes blurry, a very noticeable artifact that often leads our human evaluators to (rightly) prefer vanilla images. This lower quality may be due to the limited size and synthetic nature of the training data, as well as the small number of training parameters. This indicates that more work is needed to maintain the quality of the generated images while mitigating biases. Improvements could be achieved through more rigorous fine-tuning and the use of high-quality datasets.

\begin{figure*}
    \centering
    \includegraphics[width=1\linewidth]{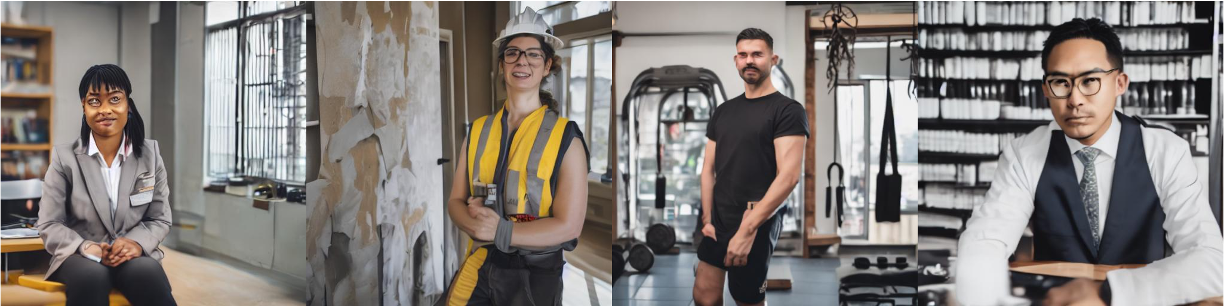}
    \caption{\textbf{Unsuccessful Generation of Face.} Althrough we achieve fairness of gender, but we lose some aesthetic details of faces.}
\end{figure*}

\section{Human Evaluation Details} 
\label{ap:human}
For the aesthetic response, we collected 29 responses, and for how well the image matches the image description, we collected 27 responses. All surveys were conducted anonymously, and the experiment was double-blinded. All the users were informed  all the collected data would be used for scientific research only. When clicking the submission button, they know the data collection is anonymous and consent to the collection.

\section{Dataset and Computing Infrastructure Explanation}

Our dataset uses the prompt list from Fair Diffusion, which includes a list of 153 occupations as shown in Appendix \ref{ap:occupation}. The experiment was conducted on NVIDIA RTX A6000 (48 GB) and Tesla V100-PCIE (32 GB) GPUs, running on Linux kernel version 6.5.0-41-generic.

\section{Randomness and Number of Algorithm Runs}

We set the random seeds for initial noise so that our results, as well as the DebiasVL results and vanilla results, remain consistent regardless of the number of runs. For Ethical Intervention and Fairness Finetuning, we perform a single run.

\section{Limitations}
Although our method has made significant strides in mitigating bias while maintaining aesthetic quality, we must acknowledge the limitations of our work.

First, we only address the case of binary gender with two bias experts. Considering sexual minorities in various contexts is a much more complex task and might need more precise bias identification methods and additional experts. We hope to explore this further in future work.

Second, our zero-shot and unsupervised prompt bias identification and hyperparameter selection are based on the statistical counts from Fair Diffusion \cite{friedrich2023FairDiffusion}, which may not be entirely accurate. Moreover, we only achieve 79\% accuracy in identifying bias from the prompt, which is not perfect.

Third, although we achieve good performance in aesthetics, our method does lose some fine-grained details on faces due to the quality of the fine-tuning image dataset. However, fortunately, our experts can be replaced with fine-tuned ones by users themselves, which can be further improved through specialized fine-tuning.

\section{Ethical Considerations}

Our work focuses on addressing social bias, specifically gender bias. Our research has a broader impact beyond scientific research. We take a significant stride across a wide range of industries and societies and our method marks a crucial step toward eliminating gender biases in text-to-image models.

However, by introducing the BiAs approach to mitigate bias, there is a risk that people might misuse the weights to generate more biased  content. Moreover, in our work, we only address the case of binary gender. We did not consider sexual minorities in various contexts, which is a much more complex task. We hope to explore this further in future work.

\end{document}